\documentclass[10pt,journal]{IEEEtran}
\usepackage{amsthm}
\usepackage{booktabs}
\usepackage{array}
\usepackage{hyperref}
\usepackage{overpic}
\usepackage{xcolor}
\usepackage{pifont}

\usepackage{amssymb}
\usepackage{graphicx}
\usepackage{epstopdf}
\usepackage[english]{babel}
\usepackage[numbers]{natbib}
\usepackage{bm}
\usepackage{float}
\usepackage{algorithm}
 \usepackage{dsfont}
\usepackage{multirow}
\usepackage{longtable}
\usepackage{subfigure}
\usepackage{supertabular}
\usepackage{algpseudocode}
\usepackage{amsmath}
\usepackage[justification=centering]{caption}

\usepackage{algorithm,float}
\makeatletter

\begin{document}

\title{Clusterability-Based Assessment of Potentially Noisy Views for Multi-View Clustering}

\author{Mudi Jiang, Jiahui Zhou, Xinying Liu, Zengyou He,  Zhikui Chen
\thanks{M. Jiang, J. Zhou, X. Liu  and Z. He are with School of Software, Dalian
University of Technology, Dalian, China.}
\thanks{Z. Chen (corresponding author) is with School of Software, Dalian University of Technology, Dalian,
China, and Key Laboratory for Ubiquitous Network and Service Software
of Liaoning Province, Dalian, China.\protect\\ Email: zkchen@dlut.edu.cn}
\thanks{Manuscript received XXXX 2026; revised XXXX, 2026.}
}
\markboth{XXX,~Vol.~XX, No.~XX, APRIL~2026}
{Shell \MakeLowercase{\textit{et al.}}: A Sample Article Using IEEEtran.cls for IEEE Journals}


\maketitle

\begin{abstract}
In multi-view clustering, the quality of different views may vary substantially, and low-quality or degraded views can impair overall clustering performance. However, existing studies mainly address this issue within the clustering process through view weighting or noise-robust optimization, while paying limited attention to data-level assessment before clustering. In this paper, we study the problem of pre-clustering noisy-view analysis in multi-view data from a clusterability perspective. To this end, we propose a Multi-View Clusterability Score (MVCS), which quantifies the strength of latent cluster-related structures in multi-view data through three complementary components: per-view structural clusterability, joint-space clusterability, and cross-view neighborhood consistency. To the best of our knowledge, this is the first clusterability score specifically designed for multi-view data. We further use it to perform potentially noisy view analysis and noisy-view detection before clustering.  Extensive experiments on real-world datasets demonstrate that noisy views can significantly degrade clustering performance, and that, compared with existing clusterability measures designed for single-view data, the proposed method more effectively supports noisy-view analysis and  detection.
\end{abstract}


\section{Introduction}
\label{1}
In recent years, multi-view clustering \cite{fang2023comprehensive} has attracted considerable attention. The objective of this problem is to exploit the complementary information contained in multiple views to obtain more stable and accurate clustering results. To address this task, numerous multi-view clustering methods have been proposed, including subspace learning–based methods, spectral clustering–based methods, and co-learning–based approaches. By integrating information from multiple views through diverse strategies, these methods have significantly advanced the development of multi-view clustering.

Building upon these efforts, several extended research directions have emerged in recent years. For instance, incomplete multi-view clustering \cite{liu2018late,yin2025incomplete} investigates how to recover and exploit multi-view information when certain views are partially missing. Robust multi-view clustering \cite{sun2024robust,zheng2025cluster,huang2023fast} focuses on enhancing model stability in the presence of noise or outliers. In addition, a growing body of research explores multi-view clustering in large-scale data scenarios \cite{Wang_2022_CVPR,yang2024methods}. Collectively, these studies have expanded the scope of multi-view clustering from multiple perspectives.

However, an important issue remains relatively underexplored. In real-world datasets, the quality of different views can vary substantially. Some views may contain significant noise or may even be irrelevant to the underlying clustering structure. When such low-quality views are directly incorporated into a multi-view clustering model, they can disrupt the structural consistency across views and consequently degrade the overall clustering performance. Most existing studies primarily focus on designing more sophisticated clustering models, while paying limited attention to whether the data itself exhibits a reliable clustering structure. If the underlying data lack stable clustering patterns, even highly complex clustering algorithms may fail to produce meaningful results.

More recently, some studies have recognized that different views may contribute unequally to multi-view clustering, and that noisy or unreliable views can harm clustering performance \cite{ye2018multi,xu2024investigating,pmlr-v209-liu23a,tang2022deep}. Most existing approaches address this issue within the clustering process through view weighting or noise-robust optimization, while paying limited attention to pre-clustering assessment of multi-view data itself. Although effective, such strategies often involve higher computational cost, more complicated optimization, and reduced flexibility in practical use due to their tight coupling with the clustering process.

Motivated by the above observation, we study the problem of pre-clustering potentially noisy view analysis in multi-view data. The objective of this task is to evaluate the structural properties of multi-view data prior to clustering, thereby supporting the analysis of views that may be detrimental to clustering performance. To this end, we propose \textbf{M}ulti-\textbf{V}iew \textbf{C}lusterability \textbf{S}core (MVCS), which quantifies the strength of latent clustering structures in multi-view data. Specifically, this score consists of three complementary structural components that capture clustering characteristics from different perspectives: single-view structure, global structure, and cross-view local neighborhood relationships. Based on this evaluation, we further perform potentially noisy view analysis under single-view perturbations.

To validate the necessity of the proposed research problem and the effectiveness of the proposed method, we conduct experiments on real-world datasets. The results show that noisy or degraded views can significantly impair clustering performance. Moreover, compared with existing clusterability measures designed for single-view data, the proposed method more effectively supports noisy-view analysis and  detection in multi-view data, thereby demonstrating its usefulness for pre-clustering data-level assessment.

The main contributions of this paper are as follows:
\begin{itemize}
    \item We formulate the problem of pre-clustering potentially noisy view analysis for multi-view clustering, aiming to analyze whether view quality may negatively affect the intrinsic clustering structure before clustering is performed.
    \item We propose MVCS,  a multi-view clusterability score which characterizes the strength of latent clustering structures in multi-view data from multiple perspectives.
    \item Extensive experiments on real-world datasets demonstrate that potentially noisy views can significantly degrade clustering performance, and verify that the proposed method can effectively identify noisy views in multi-view data.
\end{itemize}

The rest of this paper is organized as follows. Section \ref{2} reviews the related literature. Section \ref{3} presents the proposed method in detail. Section \ref{4} reports the experimental results. Finally, Section \ref{5} concludes the paper and discusses potential directions for future work.
\section{Related work}
\label{2}
Since pre-clustering analysis of potentially noisy views in multi-view data has received little attention, this section reviews the related literature from two perspectives. First, we summarize existing multi-view clustering methods, particularly those concerning view weighting and view importance. Second, because our method relies on assessing data clusterability, we also review clusterability evaluation methods for single-view data.
\subsection{Multi-View Clustering Methods}
Existing multi-view clustering approaches can generally be categorized into several groups. Graph-based methods \cite{li2021consensus, liang2019consistency} model the similarity structure of each view by constructing view-specific graphs and then exploit graph fusion or graph alignment to derive a consensus clustering result. Subspace-based methods \cite{li2019flexible} learn a shared low-dimensional representation that preserves the common information across views, thereby facilitating consistent clustering while alleviating the influence of view-specific noise. Matrix-factorization-based methods \cite{yang2020uniform,wang2018multiview} decompose multi-view data into latent low-rank factors and perform clustering based on the recovered hidden structure. Kernel-based methods \cite{liu2023contrastive} employ kernel mappings to characterize nonlinear relationships in different views and achieve clustering by integrating multiple kernels or enforcing cross-view kernel consistency. Deep learning  methods \cite{10.5555/3367243.3367449,gao2020cross,xie2020joint, xu2022self} use neural networks to learn discriminative multi-view representations and typically integrate feature learning with clustering in an end-to-end manner.

More recently, a growing body of research has recognized that different views do not contribute equally to multi-view clustering, and that some views may even degrade clustering performance due to noise, redundancy, or adverse cross-view interactions. Some studies have explicitly addressed the noisy-view problem. For example, Ye et al. \cite{ye2018multi} jointly learn view-specific clustering structures, a latent consensus clustering structure, and view weights, such that noisy views are assigned smaller weights during optimization. Xu et al. \cite{xu2024investigating} examine the detrimental effects of noisy views in self-supervised multi-view clustering and enhance robustness through a noise-aware deep optimization framework. Meanwhile, methods within the multi-view NMF framework characterize view importance through adaptive weighting. For instance, Liu et al. \cite{pmlr-v209-liu23a} learn both view-specific weights and observation-specific reconstruction weights, enabling a finer-grained assessment of view contribution and sample-level reliability. Unlike the above approaches, DSMVC \cite{tang2022deep} emphasizes the risk associated with increasing the number of views and introduces a safe module to alleviate performance deterioration when additional views are incorporated.

Although these methods have demonstrated the importance of modeling view reliability, they all incorporate noise suppression or view weighting into the clustering process and rely on joint optimization. As a result, the analysis of unreliable views is tightly coupled with specific clustering models. In contrast, our study focuses on a pre-clustering setting, where potentially noisy views are analyzed at the data level before any clustering algorithm is applied. This setting is fundamentally different from existing approaches, as it does not estimate view reliability during clustering, but instead provides a model-agnostic assessment framework that can be used independently of downstream multi-view clustering methods. Such a design offers greater flexibility in practice and makes it possible to analyze potentially noisy views without introducing additional clustering-oriented optimization.
\subsection{Clusterability Evaluation Methods}
Existing clusterability evaluation methods are primarily designed for single-view data, and can generally be divided into two categories: methods based on multimodality testing and methods based on spatial randomness testing. The former typically infers potential cluster structure by examining whether the data distribution exhibits multimodal characteristics. Among them, the Dip test \cite{cheng1998calibrating} detects multimodality by measuring the maximum deviation between the empirical distribution and the closest uniform distribution, while the Silverman test \cite{silverman1981using} is based on kernel density estimation and uses the critical bandwidth to assess the significance of modal structure in the data. The latter interprets clusterability as a deviation from spatial randomness. In particular, the Hopkins statistic \cite{dubes1987test} evaluates whether the data tend to be aggregated by comparing nearest-neighbor distances between real samples and randomly generated samples, although its results depend on random sampling and may suffer from limited stability and applicability in high-dimensional settings. In contrast, PHI \cite{diallo2023deciphering} jointly characterizes compactness and separability among samples through a knowledge graph, and thus evaluates structural homogeneity in a deterministic manner. Overall, these methods all assess clusterability based on a single-view data representation, and have not been specifically designed for quantitative comparison of structural differences across views in multi-view scenarios.

In this work, we draw inspiration from the Silverman test, particularly its idea of characterizing the persistence of multimodal structure through the critical bandwidth. However, unlike the original Silverman test, which relies on bootstrap-based significance testing to make a binary decision on unimodality or multimodality, our method does not aim to directly determine whether a view is “clusterable” or not. Instead, we directly adopt the critical bandwidth as a continuous structural indicator that reflects the persistence of multimodal patterns in the data, and further transform it into a normalized score. In this way, the original hypothesis testing framework is converted into a quantitative evaluation scheme that preserves richer structural information beyond a binary outcome and enables fine-grained comparison across different views in multi-view data.
\section{Method}
\label{3}
This section provides a detailed description of the proposed multi-view data clusterability score, which consists of three components: per-view structural clusterability, joint-space clusterability, and cross-view neighborhood consistency.
\subsection{Per-view Structural Clusterability}
Consider a multi-view dataset consisting of $N$ instances observed from $V$ distinct views $X=\{X^{(v)}\}_{v=1}^{V}$,  $X^{(v)} \in \mathbb{R}^{N \times d_v}$, where $d_v$ denotes the dimensionality of the 
$v$-th view. Our objective is to quantify whether each individual view intrinsically exhibits cluster-indicative structural patterns.

Since the geometric distribution of high-dimensional data is difficult to analyze directly, we follow the intuition of the Silverman test and derive a one-dimensional structural representation for each view, so that multimodal structure can be analyzed through kernel density estimation and critical bandwidth. Specifically, after standardization, each view is projected onto its principal direction of variation, yielding a one-dimensional sample set $Y^{(v)}=\{y^{(v)}_1,y^{(v)}_2,\dots,y^{(v)}_N\}$. To characterize the structural separability of
$Y^{(v)}$, a Gaussian kernel density estimator (KDE) is constructed with bandwidth $h > 0$:
\begin{equation}
    \hat{f}_h(x)=\frac{1}{Nh}\sum_{i=1}^{N} K\!\left(\frac{x-y^{(v)}_i}{h}\right),
\end{equation}
where $K(\cdot)$ denotes the standard Gaussian kernel.
The bandwidth parameter $h$ controls the smoothness of the density estimate. As $h$ increases, the density curve becomes progressively smoother and the number of local maxima decreases monotonically. Let $M_\text{peaks}(h)$ denote the number of modes of $\hat{f}_h(x)$. The critical bandwidth corresponding to unimodality is defined as
\begin{equation}
    h^{(v)}_{\mathrm{crit}}=\inf\{h>0: M_{\text{peaks}}(h)\le 1\},
\end{equation}
where the infimum selects the smallest bandwidth at which the density estimate becomes unimodal, corresponding to the minimal smoothing strength required to eliminate multi-modality. The critical bandwidth admits a clear structural interpretation. A nearly unimodal distribution requires only slight smoothing, resulting in a small $h^{(v)}_{\mathrm{crit}}$. In contrast, when multiple stable modes induced by latent cluster structures exist, substantially stronger smoothing is needed to merge them, leading to a larger $h^{(v)}_{\mathrm{crit}}$. Therefore, $h^{(v)}_{\mathrm{crit}}$ serves as a continuous indicator reflecting the persistence of multi-modal structures within the view.

Unlike the original Silverman hypothesis testing framework, which aims to test the null hypothesis of at most 
$k$ modes via bootstrap-based significance estimation, our objective is fundamentally different. We do not seek a binary decision regarding unimodality. Instead, we reinterpret $h^{(v)}_{\mathrm{crit}}$ as a continuous structural measure that quantifies the persistence of multi-modality. The original procedure yields a statistical decision outcome, whereas our aim is to construct a quantitative scoring function that captures the relative strength of cluster-indicative structures. By directly adopting $h^{(v)}_{\mathrm{crit}}$ and transforming it into a normalized score, we preserve richer structural information beyond a binary conclusion, thereby enabling fine-grained comparison across views within a unified evaluation framework.

To ensure scale invariance and comparability across views, we normalize the critical bandwidth by the dispersion of the projected data. Let $\sigma^{(v)}_y$ denote the standard deviation of $Y^{(v)}$. The per-view structural separability score is defined as
\begin{equation}
    s(v)=1-\exp\!\left(-\frac{h^{(v)}_{\mathrm{crit}}}{\tau \sigma^{(v)}_y}\right),
\end{equation}
where $\tau > 0$ controls the sensitivity of the transformation. This monotonic mapping preserves the relative ordering while constraining the score to the interval $[0,1]$. Finally, the per-view clusterability separability component of the proposed framework is obtained by averaging across all views:
\begin{equation}
    S_{\mathrm{pv}}=\frac{1}{V}\sum_{v=1}^{V} s(v).
\end{equation}
This term quantifies the intrinsic cluster-indicative strength independently contained in each view and constitutes the first component of the overall multi-view clusterability score.
\subsection{Joint-Space Clusterability}
While the previous component evaluates the structural separability of each view independently, multi-view data may also exhibit additional cluster-related information after combining features from different views. Even when individual views are not strongly separable on their own, their joint representation may still present a clearer overall structure. Therefore, in addition to per-view analysis, it is useful to assess clusterability in the joint feature space. To this end, we construct a unified representation by concatenating the standardized features from all views:
\begin{equation}
    Z_i=[x_i^{(1)};x_i^{(2)};\dots;x_i^{(V)}],
\end{equation}
where $Z_i$ denotes the aggregated feature vector of the $i$-th instance. Here we intentionally adopt simple feature concatenation rather than more complex consensus representation learning, because our goal is to assess the intrinsic clusterability of the data before clustering, rather than to improve the representation through an additional learning procedure.

Similar to the per-view analysis, the concatenated representation is projected onto its principal variation direction to obtain a one-dimensional sample set
\begin{equation}
    Y^{(c)}=\{y^{(c)}_1,y^{(c)}_2,\dots,y^{(c)}_N\}.
\end{equation}
The same density morphology analysis is then applied. Specifically, a Gaussian kernel density estimator is constructed and the critical bandwidth $h_{\mathrm{crit}}^{(c)}$ required to achieve unimodality is computed. Following the same normalization strategy, the joint-space clusterability score is defined as
\begin{equation}
    S_{\mathrm{joint}} =
    1-\exp\!\left(
    -\frac{h_{\mathrm{crit}}^{(c)}}{\tau \sigma^{(c)}_y}
    \right),
\end{equation}
where $\sigma^{(c)}_y$ denotes the standard deviation of $Y^{(c)}$.

\subsection{Cross-View Neighborhood Consistency}
The previous two components mainly characterize the structural properties of the data distribution. However, cluster-related organization is also reflected in the stability of local neighborhoods. Intuitively, if samples that are close to each other in one view tend to remain neighbors in other views, then the local geometry is relatively consistent across views, which may provide useful support for clusterability assessment. In contrast, if neighborhood relationships vary substantially across views, the structural organization of the data becomes less stable.

To quantify such local structural agreement, we analyze neighborhood consistency across views. For each instance $x_i^{(v)}$ in the $v$-th view, we denote by $\mathcal{N}_i^{(v)}$ the set of indices corresponding to its $k$ nearest neighbors within the same view. To accelerate nearest-neighbor search in high-dimensional spaces, the KNN computation is implemented using an efficient approximate nearest neighbor algorithm based on the FAISS library \cite{11202651}.

Given two views $v$ and $u$, the neighborhood agreement for instance $i$ between these views is measured by the overlap ratio of their neighbor sets:
\begin{equation}
a_i^{(v,u)} =
\frac{
|\mathcal{N}_i^{(v)} \cap \mathcal{N}_i^{(u)}|
}{k}.
\end{equation}
This quantity represents the proportion of neighbors that remain consistent across the two views. Larger values indicate that the local structure around the instance is more stably preserved across views.
To measure neighborhood stability across all views, we average the agreement scores over all view pairs:
\begin{equation}
a_i =
\frac{2}{V(V-1)}
\sum_{v<u} a_i^{(v,u)}.
\end{equation}
Finally, the overall multi-view neighborhood consistency score is obtained by averaging across all instances:
\begin{equation}
S_{\mathrm{nbr}} =
\frac{1}{N}
\sum_{i=1}^{N} a_i.
\end{equation}
This component reflects the stability of local geometric relationships across views. When multiple views preserve similar neighborhood structures, the score becomes large, indicating stronger cross-view local consistency. In contrast, inconsistent neighborhood relationships lead to lower scores, suggesting that the local structure is less stable across views.

To obtain a unified evaluation, we first combine the above three components into a raw multi-view clusterability score:
\begin{equation}
S_{\mathrm{raw}} =
\alpha S_{\mathrm{pv}} +
\beta S_{\mathrm{joint}} +
\gamma S_{\mathrm{nbr}},
\end{equation}
where $\alpha$, $\beta$, and $\gamma$ are non-negative weights satisfying $\alpha + \beta + \gamma = 1$.
 The final calibrated clusterability score is then defined as
\begin{equation}
S = 1 - \exp(-S_{\mathrm{raw}} / \eta),
\end{equation}
where $\eta$ is a scaling parameter.
\section{Experiments}
\label{4}

In this section, we present a series of experiments conducted on real-world datasets to assess the effectiveness of the proposed method. Specifically, the experiments are designed to address the following research questions (RQs):
\begin{itemize}
\item \textbf{RQ1:} Do noisy views have a significant impact on multi-view clustering performance?
\item \textbf{RQ2:} Can the proposed score effectively characterize the degradation of clusterability in multi-view data as noisy views are introduced?
\item \textbf{RQ3:} Is the proposed  score effective under controlled view perturbation?
\item \textbf{RQ4:} Can the proposed method more effectively identify potentially noisy views?
\end{itemize}
Here, \textbf{RQ1} aims to validate the necessity of the proposed problem from a data-centric perspective by examining whether noisy views indeed degrade multi-view clustering performance. \textbf{RQ2} evaluates whether the proposed score can effectively reflect the degradation of cluster-related structure in multi-view data as the number of noisy views increases. \textbf{RQ3} investigates whether the proposed score can effectively characterize structural changes under controlled view-level perturbations. \textbf{RQ4} focuses on the practical noisy-view detection setting, namely, whether the proposed method can automatically identify potentially noisy views from a given set of multi-view data.

All experiments were carried out on a workstation equipped with an Intel(R) Core(TM) i7-10700F CPU at 2.90 GHz, 16 GB RAM, and an NVIDIA GeForce RTX 1660 GPU with 6 GB of memory.
\subsection{Experimental Setup}
\begin{table}[htbp]
	\caption{The summary statistics on datasets used in the performance evaluation.}
	\label{dataset}
	\centering
        \setlength{\tabcolsep}{0.62mm}{
	\begin{tabular}{ccccccc}
		\toprule
		Dataset &  \#Samples &\#Features& \#Clusters &\#Views  \\
		\midrule
		Mfeat &  2000  &216/76/64/6/240/47 &10 &  6         \\
            ORL &  400 &4096/3304/6750 &40  &  3         \\
            MSRCV1 &  210 &1302/48/512/100/256/210 &7  & 6          \\
		COIL &  1440  &1024/3304/6750 & 20&  3       \\
		Caltech-5V &  1400  &40/254/1984/512/928 &7 &  5        \\
        20NGs &  500  &2000/2000/2000 &5 &  3        \\
		\bottomrule
	\end{tabular}}
\end{table}

\begin{figure*}[htbp]
	\includegraphics[scale=0.6]{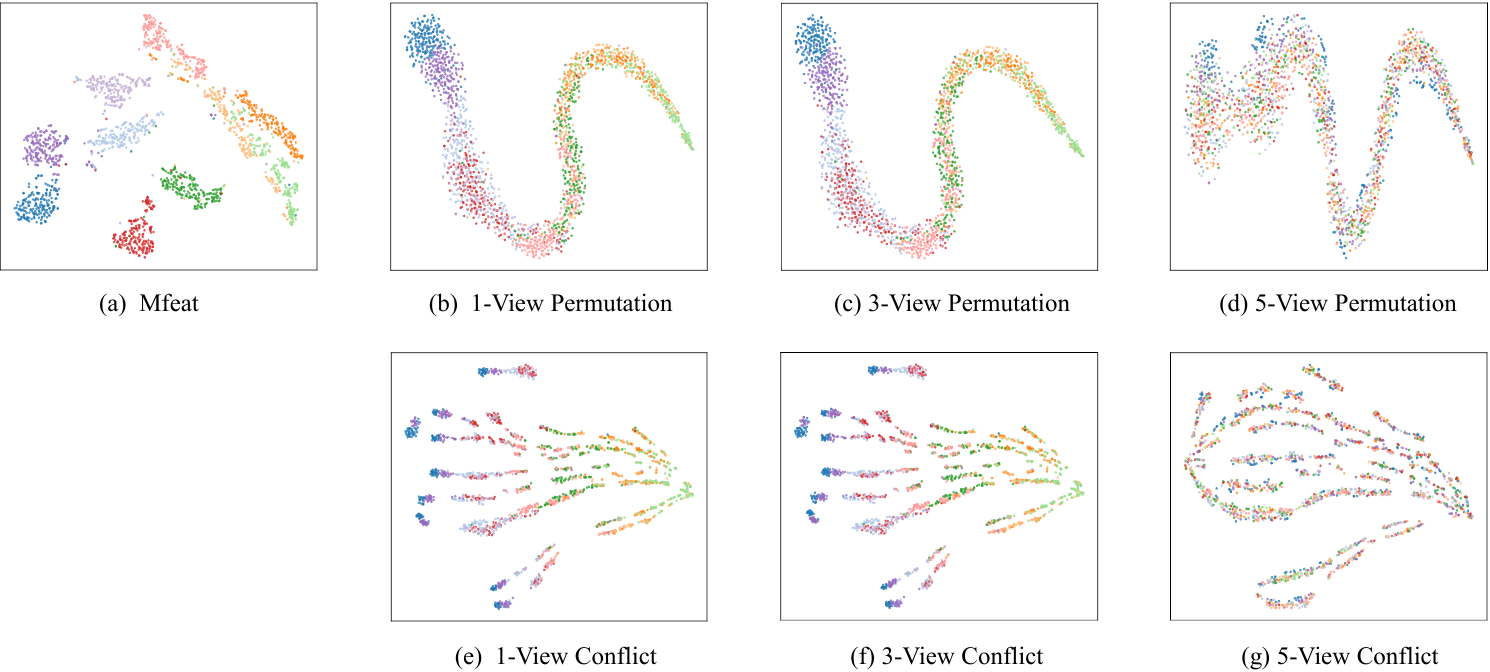}  
	\centering
	\caption{t-SNE visualization  with varying numbers of noisy views on the Mfeat dataset.}  
	\label{tsne_mfeat}   
\end{figure*}

\begin{figure*}[htbp]
	\includegraphics[scale=0.6]{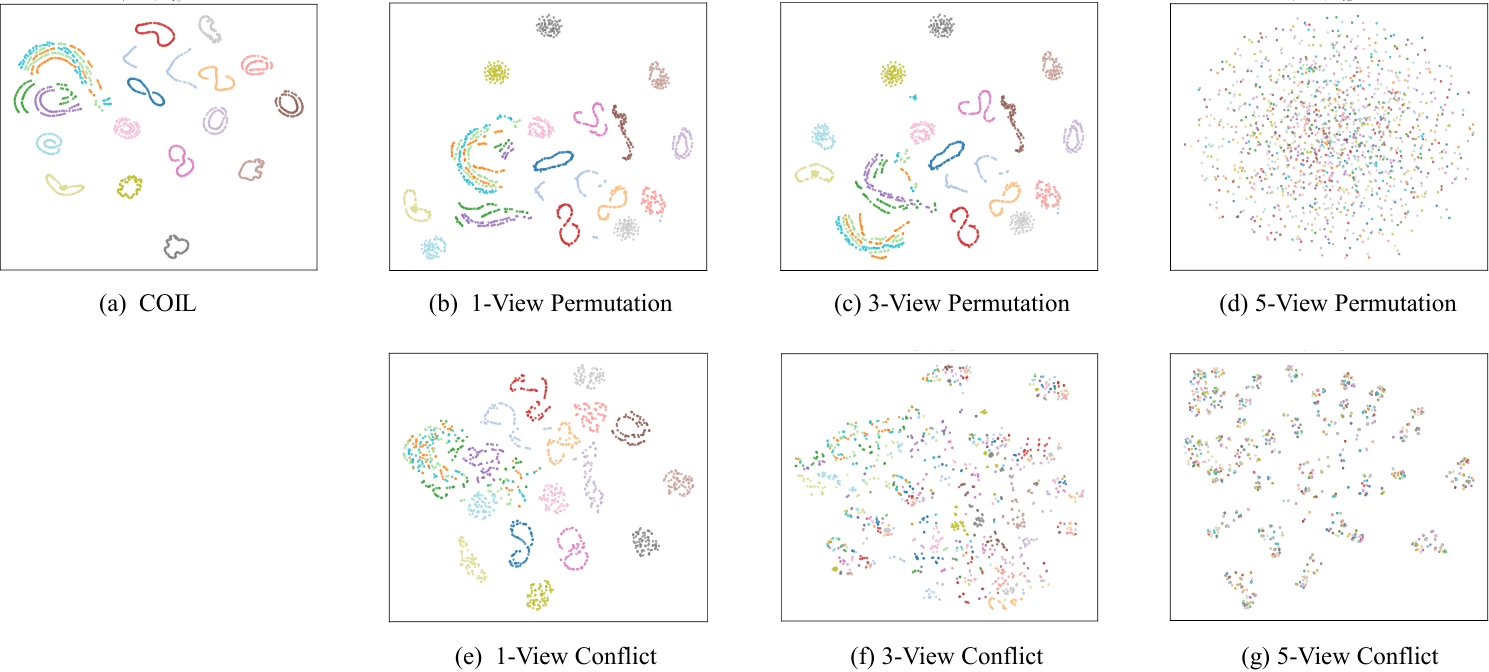}  
	\centering
	\caption{t-SNE visualization  with varying numbers of noisy views on the COIL dataset.}  
	\label{tsne_COIL}   
\end{figure*}

\begin{figure*}[htbp]
	\includegraphics[scale=0.55]{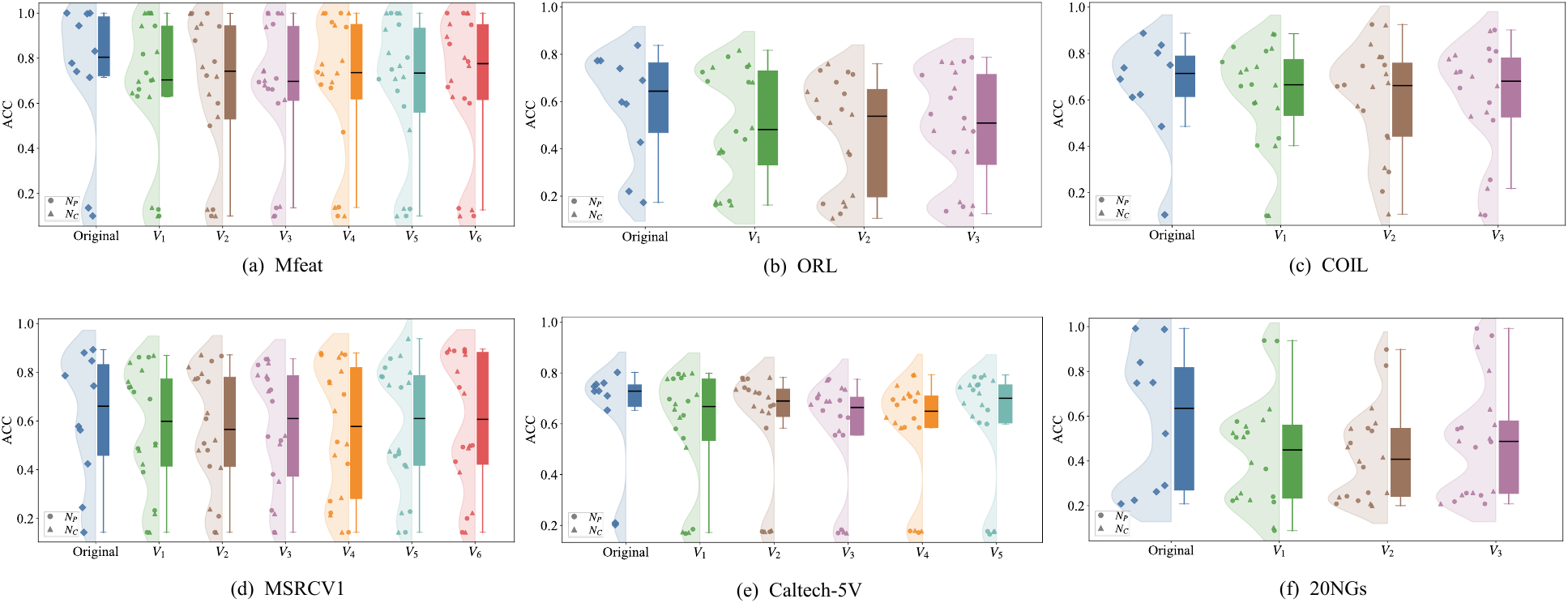}  
	\centering
	\caption{Clustering accuracy (ACC) of ten multi-view clustering algorithms when each view is individually replaced by noisy views under two corruption strategies across different datasets.}  
	\label{box}   
\end{figure*}
\begin{figure*}[htbp]
	\includegraphics[scale=0.65]{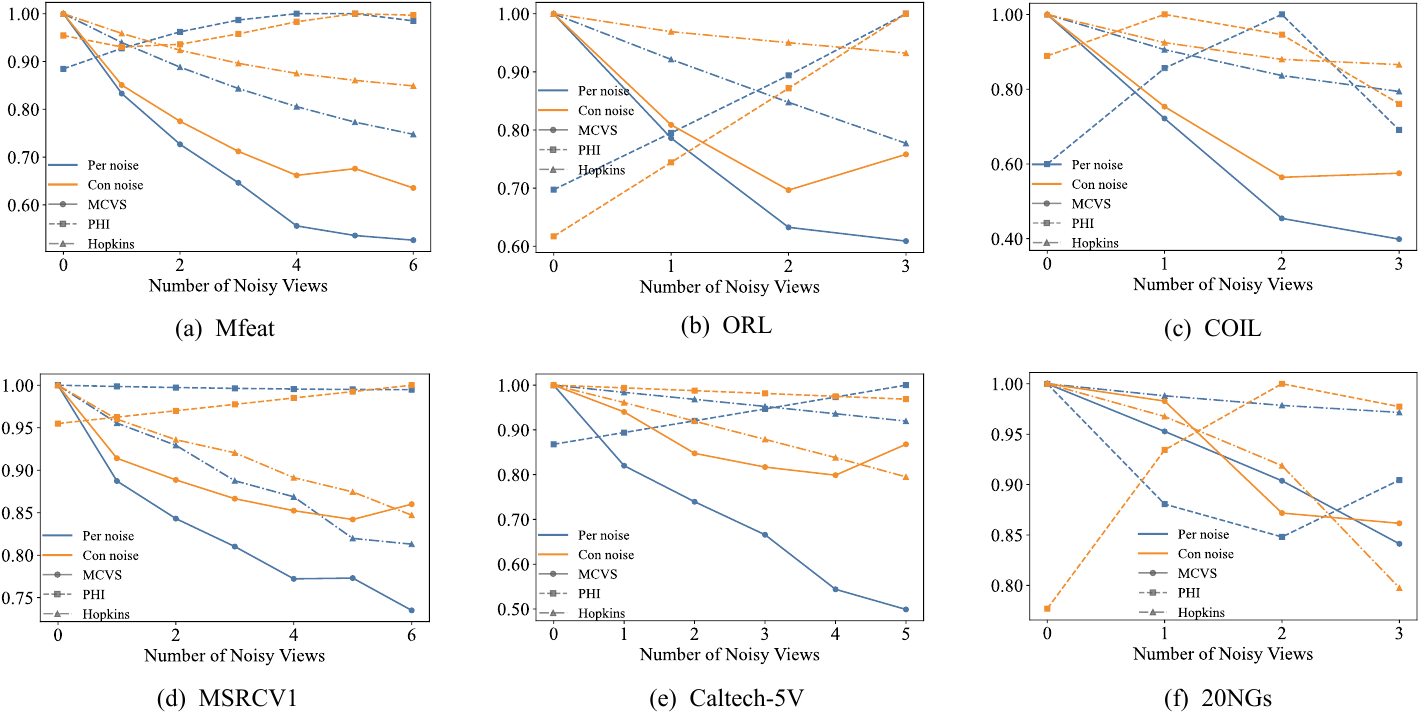}  
	\centering
	\caption{Comparison of clusterability  metrics under increasing numbers of noisy views across datasets. The vertical axis shows normalized scores. Each score series is normalized by dividing by its maximum value.}  
    \label{zx}  
\end{figure*}
Six datasets are used to evaluate the proposed method: Mfeat \footnote{https://archive.ics.uci.edu/dataset/72/multiple+features}, ORL \footnote{http://www.uk.research.att.com/facedatabase.html}, COIL \footnote{http://www.cs.columbia.edu/CAVE/software/softlib}, MSRCV1 \footnote{https://www.microsoft.com/en-us/research/project/image-understanding}, Caltech-5V \cite{xu2022multi}, and 20NGs \cite{HU2020107101}, the detailed characteristics of each dataset are summarized in Table \ref{dataset}. To simulate noisy views, we adopt two corruption strategies. 
(1) Permutation-based noise (Per): the sample order of each feature column is randomly shuffled, which preserves the marginal distribution of each feature while destroying the underlying clustering structure. 
(2) Conflict-based noise (Con): class-wise feature distributions are first estimated from the original data, and all samples are reassigned to different classes with features generated from the corresponding class distributions, thereby introducing structural conflicts with the true clustering structure. 
In our implementation, the parameters $\alpha$, $\beta$, and $\gamma$ are set to 0.2, 0.2, and 0.6, respectively.

To further examine the impact of noisy views on existing multi-view clustering algorithms, we include ten representative baselines: CHOC \cite{you2024consider}, CGL \cite{li2021consensus}, MFLVC \cite{xu2022multi}, TNNF \cite{YAN2026112969}, MCPL \cite{cai2024multi}, TDCLM \cite{WU2025121575}, MCMF \cite{10855587}, 3MC \cite{chen2025multi}, DCMVC \cite{LI2025129889}, and SSLNMVC \cite{10891409}.

\subsection{Investigation of RQ1}
To investigate the impact of noisy views on multi-view clustering, we first provide a visual analysis using the Mfeat and COIL datasets. Specifically, we apply t-SNE to the concatenated features of all views to visualize the overall data distribution. For each dataset, we compare the original data with scenarios in which the first 1, 3, and 5 views are replaced by noisy versions generated using the two corruption strategies described above. As shown in Figs. \ref{tsne_mfeat} and \ref{tsne_COIL}, even replacing a single view noticeably changes the distribution pattern and weakens cluster separability. As the number of noisy views increases, the cluster structure becomes progressively more mixed and less distinguishable, suggesting that noisy views can substantially disrupt the overall structure of multi-view data.

To further quantify this effect, Fig. \ref{box} reports the clustering accuracy (ACC) of ten representative multi-view clustering algorithms when one view at a time is replaced by a noisy version generated by the two corruption strategies. Since both noisy-view construction and clustering optimization may involve randomness, each algorithm was repeated 10 times under different random seeds for every dataset, noise type, and replaced-view setting. For each setting, the same noisy view realization was shared by all algorithms to ensure fair comparison. We report the mean ACC of each algorithm over the 10 runs, and the boxplots summarize these values across the ten algorithms under the two noise settings.

It can be seen that replacing any single view with a noisy counterpart generally leads to a noticeable decrease in clustering accuracy compared with the original data. This trend is consistently observed across different datasets, indicating that even a single noisy view may substantially impair the performance of multi-view clustering methods and revealing their sensitivity to low-quality or noisy views.
\subsection{Investigation of RQ2}

To evaluate the effectiveness of the proposed clusterability score for multi-view data, we analyze how different scores respond to increasing numbers of noisy views. As shown in Fig. \ref{zx}, each dataset is associated with a line plot where the horizontal axis represents the number of noisy views and the vertical axis denotes the normalized clusterability score. We compare  the proposed MVCS with two classical clusterability measures originally designed for single-view data, i.e., PHI and the Hopkins statistic. Since PHI indicates clearer cluster structures with smaller values, we transform it using $1-x$ before normalization so that larger values consistently correspond to stronger clusterability for all metrics.

Several observations can be drawn from the results. First, PHI often fails to reflect the degradation of cluster-related structures in multi-view data, particularly under the conflict-based noise (Con). On most datasets except Caltech-5V, PHI even exhibits an opposite trend, where the score increases as more noisy views are introduced, falsely suggesting stronger clusterability.

Second, both MVCS and Hopkins generally respond in the expected direction as the number of noisy views increases. However, their behaviors differ. The Hopkins statistic tends to change relatively smoothly as noise accumulates, whereas MVCS is more sensitive to the presence of noisy views. Even introducing a single noisy view typically leads to a noticeable decrease in the MVCS score. This behavior indicates that MVCS can more clearly capture the structural changes caused by view degradation, which is useful for controlled view-degradation analysis and further supports its application to potentially noisy view analysis in multi-view data (see Section \ref{4.4} and \ref{4.5}).
\subsection{Investigation of RQ3}
\label{4.4}
\begin{table}[htbp]
\centering
\scriptsize
\setlength{\tabcolsep}{7pt}
\renewcommand{\arraystretch}{0.83}

\caption{Response consistency under single-view perturbation settings, larger values indicate stronger clusterability.}
\label{tab:replace_drop_alpha}

\begin{tabular}{c c c c c c}
\toprule

\small Method & \small Dataset & \small View & \small Drop & \small Per & \small Con \\

\midrule

\multirow{26}{*}{\small Ours}
& \multirow{6}{*}{\shortstack[c]{\small MSRCV1\\{\small MVCS=0.713}}}
& 1 & \underline{0.722} & 0.600 & 0.652 \\
& & 2 & 0.692 & 0.642 & 0.658 \\
& & 3 & 0.699 & 0.612 & 0.649 \\
& & 4 & 0.667 & 0.627 & 0.646 \\
& & 5 & 0.683 & 0.634 & 0.646 \\
& & 6 & 0.706 & 0.680 & 0.659 \\

\cmidrule(lr){2-6}
& \multirow{6}{*}{\shortstack[c]{\small Mfeat\\{\small MVCS=0.669}}}
& 1 & 0.586 & 0.527 & 0.533 \\
& & 2 & 0.664 & 0.586 & 0.608 \\
& & 3 & 0.620 & 0.527 & 0.514 \\
& & 4 & 0.666 & 0.558 & 0.614 \\
& & 5 & 0.619 & 0.564 & 0.558 \\
& & 6 & 0.646 & 0.581 & 0.586 \\

\cmidrule(lr){2-6}
& \multirow{5}{*}{\shortstack[c]{\small Caltech-5V\\{\small MVCS=0.607}}}
& 1 & \underline{0.699} & 0.547 & \underline{0.609} \\
& & 2 & 0.593 & 0.544 & \underline{0.670} \\
& & 3 & \underline{0.726} & 0.463 & 0.552 \\
& & 4 & 0.597 & 0.476 & 0.534 \\
& & 5 & 0.557 & 0.457 & 0.487 \\

\cmidrule(lr){2-6}
& \multirow{3}{*}{\shortstack[c]{\small ORL\\{\small MVCS=0.726}}}
& 1 & 0.718 & 0.584 & 0.559 \\
& & 2 & \underline{0.743} & 0.570 & 0.631 \\
& & 3 & 0.725 & 0.557 & 0.570 \\

\cmidrule(lr){2-6}
& \multirow{3}{*}{\shortstack[c]{\small COIL\\{\small MVCS=0.866}}}
& 1 & 0.827 & 0.593 & 0.619 \\
& & 2 & \underline{0.938} & 0.691 & 0.689 \\
& & 3 & 0.857 & 0.590 & 0.649 \\

\cmidrule(lr){2-6}
& \multirow{3}{*}{\shortstack[c]{\small 20NGs\\{\small MVCS=0.790}}}
& 1 & \underline{0.826} & \underline{0.810} & 0.787 \\
& & 2 & 0.763 & 0.750 & \underline{0.800} \\
& & 3 & 0.763 & 0.698 & 0.743 \\

\cmidrule(lr){2-6}
& \small Correct Responses &  & 20/26 & 25/26 & 23/26 \\
\midrule

\multirow{26}{*}{\small Hopkins}
& \multirow{6}{*}{\shortstack[c]{\small MSRCV1\\{\small Hopkins=0.813}}}
& 1 & 0.807 & \underline{0.815} & 0.811 \\
& & 2 & \underline{0.818} & \underline{0.819} & 0.810 \\
& & 3 & 0.810 & 0.810 & 0.\underline{815} \\
& & 4 & 0.810 & 0.808 & \underline{0.818} \\
& & 5 & 0.691 & 0.664 & 0.702 \\
& & 6 & \underline{0.813} & 0.812 & 0.805 \\

\cmidrule(lr){2-6}
& \multirow{6}{*}{\shortstack[c]{\small Mfeat\\{\small Hopkins=0.813}}}
& 1 & 0.773 & 0.615 & 0.697 \\
& & 2 & 0.812 & 0.812 & \underline{0.813} \\
& & 3 & \underline{0.813} & 0.810 & \underline{0.814} \\
& & 4 & \underline{0.839} & 0.774 & 0.770 \\
& & 5 & 0.812 & 0.811 & 0.812 \\
& & 6 & \underline{0.820} & 0.760 & 0.768 \\

\cmidrule(lr){2-6}
& \multirow{5}{*}{\shortstack[c]{\small Caltech-5V\\{\small Hopkins=0.905}}}
& 1 & \underline{0.905} & \underline{0.906} & \underline{0.907} \\
& & 2 & 0.775 & 0.833 & 0.721 \\
& & 3 & \underline{0.905} & \underline{0.905} & \underline{0.905} \\
& & 4 & \underline{0.907} & \underline{0.906} & \underline{0.905} \\
& & 5 & 0.904 & 0.904 & \underline{0.907} \\

\cmidrule(lr){2-6}
& \multirow{3}{*}{\shortstack[c]{\small ORL\\{\small Hopkins=0.734}}}
& 1 & 0.699 & 0.572 & 0.687 \\
& & 2 & 0.734 & 0.736 & 0.732 \\
& & 3 & 0.735 & 0.734 & 0.728 \\

\cmidrule(lr){2-6}
& \multirow{3}{*}{\shortstack[c]{\small COIL\\{\small Hopkins=0.815}}}
& 1 & \underline{0.815} & \underline{0.816} & \underline{0.815} \\
& & 2 & \underline{0.883} & 0.680 & 0.707 \\
& & 3 & 0.803 & 0.716 & 0.739 \\

\cmidrule(lr){2-6}
& \multirow{3}{*}{\shortstack[c]{\small 20NGs\\{\small Hopkins=0.943}}}
& 1 & \underline{0.951} & 0.929 & 0.908 \\
& & 2 & \underline{0.943} & 0.937 & 0.900 \\
& & 3 & \underline{0.945} & 0.918 & 0.940 \\

\cmidrule(lr){2-6}
& \small Correct Responses &  & 13/26 & 20/26 & 17/26 \\
\midrule

\multirow{26}{*}{\small PHI}
& \multirow{6}{*}{\shortstack[c]{\small MSRCV1\\{\small PHI=0.717}}}
& 1 & \underline{0.717} & \underline{0.717} & \underline{0.717} \\
& & 2 & \underline{0.717} & \underline{0.717} & \underline{0.717} \\
& & 3 & \underline{0.717} & \underline{0.717} & \underline{0.717} \\
& & 4 & \underline{0.717} & \underline{0.717} & \underline{0.717} \\
& & 5 & 0.634 & 0.711 & \underline{0.751} \\
& & 6 & \underline{0.717} & \underline{0.717} & \underline{0.717} \\

\cmidrule(lr){2-6}
& \multirow{6}{*}{\shortstack[c]{\small Mfeat\\{\small PHI=0.359}}}
& 1 & 0.239 & \underline{0.381} & 0.351 \\
& & 2 & \underline{0.359} & \underline{0.359} & \underline{0.359} \\
& & 3 & \underline{0.359} & \underline{0.359} & \underline{0.359} \\
& & 4 & \underline{0.790} & \underline{0.469} & \underline{0.359} \\
& & 5 & \underline{0.359} & \underline{0.359} & 0.356 \\
& & 6 & 0.339 & 0.332 & 0.318 \\

\cmidrule(lr){2-6}
& \multirow{5}{*}{\shortstack[c]{\small Caltech-5V\\{\small PHI=0.731}}}
& 1 & \underline{0.731} & \underline{0.731} & \underline{0.731} \\
& & 2 & 0.621 & \underline{0.841} & 0.708 \\
& & 3 & \underline{0.731} & \underline{0.731} & \underline{0.731} \\
& & 4 & \underline{0.731} & \underline{0.731} & \underline{0.731} \\
& & 5 & \underline{0.731} & \underline{0.731} & \underline{0.731} \\

\cmidrule(lr){2-6}
& \multirow{3}{*}{\shortstack[c]{\small ORL\\{\small  PHI=0.443}}}
& 1 & 0.382 & \underline{0.629} & \underline{0.717} \\
& & 2 & 0.442 & \underline{0.443} & 0.442 \\
& & 3 & \underline{0.443} & \underline{0.443} & \underline{0.443} \\

\cmidrule(lr){2-6}
& \multirow{3}{*}{\shortstack[c]{\small COIL\\{\small  PHI=0.414}}}
& 1 & 0.405 & 0.404 & \underline{0.429} \\
& & 2 & 0.350 & \underline{0.608} & \underline{0.483} \\
& & 3 & \underline{0.429} & \underline{0.760} & \underline{0.484} \\

\cmidrule(lr){2-6}
& \multirow{3}{*}{\shortstack[c]{\small 20NGs\\{\small  PHI=0.446}}}
& 1 & 0.424 & 0.424 & 0.423 \\
& & 2 & 0.307 & \underline{0.668} & 0.132 \\
& & 3 & \underline{0.446} & 0.445 & \underline{0.446} \\

\cmidrule(lr){2-6}
& \small Correct Responses &  & 10/26 & 5/26 & 7/26 \\

\bottomrule
\end{tabular}
\end{table}
\begin{figure*}[htbp]
	\includegraphics[scale=0.6]{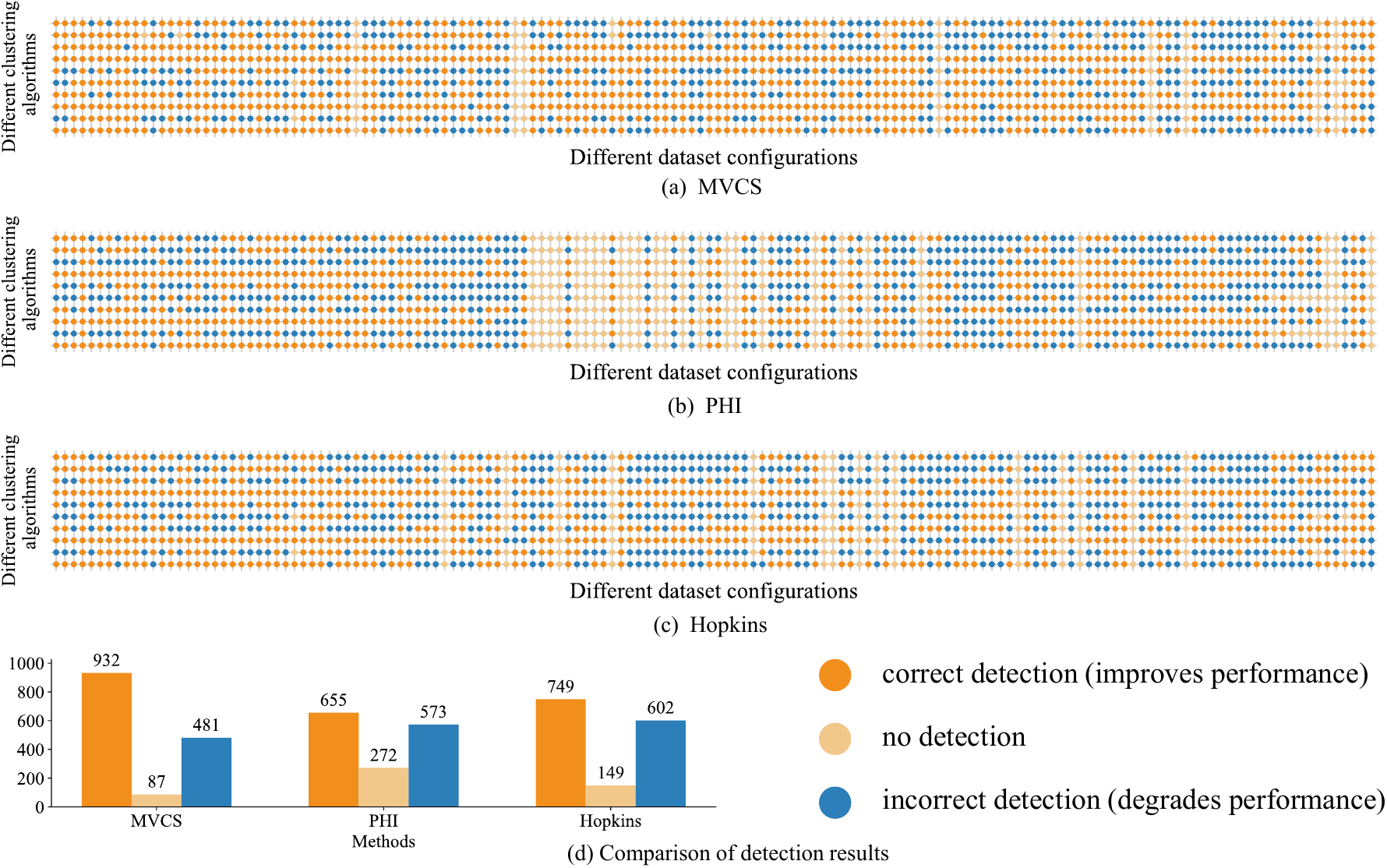}  
	\centering
	\caption{Noisy-view detection results of different clusterability scores. Subfigures (a), (b), and (c) correspond to MVCS, PHI, and Hopkins statistic, respectively. Each row represents a multi-view clustering algorithm, and each column represents a noisy data configuration. Orange dots indicate that removing the detected noisy view under the current configuration improves clustering accuracy (ACC), while blue dots indicate the opposite. Light orange dots indicate that the corresponding clusterability score does not detect any noisy view in the data. Subfigure (d) summarizes and compares the overall performance of different scores.}  
	\label{qp}   
\end{figure*}
In Table \ref{tab:replace_drop_alpha}, we evaluate the response of the proposed clusterability score to single-view removal and controlled noisy replacement, and compare it with PHI and the Hopkins statistic. The “Drop” column reports the clusterability score obtained after removing a single view from the original multi-view data. For an informative view, its removal is expected to decrease the score; if the score instead increases, the method has incorrectly treated the original informative view as harmful, indicating a false positive. The “Per” and “Con” columns report the scores obtained after replacing a single view with a permuted noisy view and a structurally conflicting noisy view, respectively. A decrease in the score after replacement indicates that the method can capture the structural degradation caused by the noisy view, thereby supporting its usefulness for potentially noisy view analysis. Several important observations can be drawn from these results:

Compared with PHI, the proposed method exhibits a stronger response under both types of noisy views, while also producing fewer false positives. This is mainly because PHI is designed to deterministically measure the overall homogeneity of data structure. As such, it is better suited to evaluating clusterability from the perspective of global sample relationships, rather than specifically characterizing the structural disruption caused by the degradation of an individual view. Consequently, when a particular view is replaced by noise, PHI is relatively less sensitive to this type of view-level perturbation.

Compared with the Hopkins statistic, the proposed method likewise demonstrates a stronger response to noisy views, with a particularly clear advantage in terms of false positives. Although the Hopkins statistic can also respond to noisy views to some extent, making the difference in responsiveness less pronounced, it is essentially a spatial randomness test that relies on random sampling to compare nearest-neighbor relationships between real and randomly generated samples. Its results are therefore less stable and more susceptible to sampling variation and the complexity of the data distribution, which in turn leads to more false positives.

Overall, these results indicate that the proposed score is more appropriate for controlled single-view perturbation analysis than PHI and the Hopkins statistic. It responds more clearly to noisy-view replacement while producing fewer false positives under view removal, suggesting that it can more effectively reflect the influence of view-level degradation on the overall cluster-related structure of multi-view data.
\subsection{Investigation of RQ4}
\label{4.5}

This section analyzes whether different methods can accurately identify noisy views in multi-view data contaminated by noise. Specifically, for each of the two noise types, we replace one or two views with noisy views on the six datasets introduced earlier, resulting in a total of 150 different data configurations. The rule for noisy-view detection is defined as follows. We remove each view individually and compute the corresponding change in the clusterability score. If $Score(\hat{X}) - Score(X) \geq 0$ (or $\leq 0$  for PHI), where $X$ denotes the original multi-view data and $\hat{X}$ represents the data after removing a candidate view, this indicates that removing the corresponding view leads to a higher overall clusterability score, and the view is thus regarded as a potential noisy view. If multiple potential noisy views are identified, the one with the largest score change is selected.

Based on this, we further apply ten different clustering algorithms to the data before and after removing the detected noisy view. If the clustering accuracy (ACC) improves after removing the noisy view, the detection is regarded as successful; otherwise, it is considered a failed detection. The detailed results are presented in Fig. \ref{qp}, which reports the performance of the three clusterability scores as well as their comparison. It can be observed that the proposed MVCS achieves the highest number of correct detections and the lowest number of incorrect detections among all methods, demonstrating the effectiveness of the proposed clusterability-based framework for noisy-view detection.

\section{Conclusion}
\label{5}
In this paper, we studied the problem of pre-clustering noisy-view detection in multi-view data from a clusterability perspective. To this end, we proposed a Multi-View Clusterability Score (MVCS), which integrates per-view structural clusterability, joint-space clusterability, and cross-view neighborhood consistency. Different from existing methods that estimate view reliability during clustering, the proposed method provides a model-agnostic, pre-clustering assessment of multi-view data. Experimental results on real-world datasets demonstrated that the proposed method more effectively supports noisy-view analysis and  detection than classical single-view clusterability measures.

In future work, we will investigate more refined view-level assessment strategies and explore stronger connections between clusterability analysis and specific multi-view clustering methods. In particular, it is worth studying whether the proposed score can provide useful guidance for view weighting, view selection, and robust fusion in downstream clustering models.
\appendices

\section*{Acknowledgments}
This work has been supported by the National Natural Science Foundation of China under Grant Nos. 62476038, and 62472064. 

\small
\bibliographystyle{IEEEtran}
\bibliography{refj}

\begin{IEEEbiography}[{\includegraphics[width=1in,height=1.25in,clip,keepaspectratio]{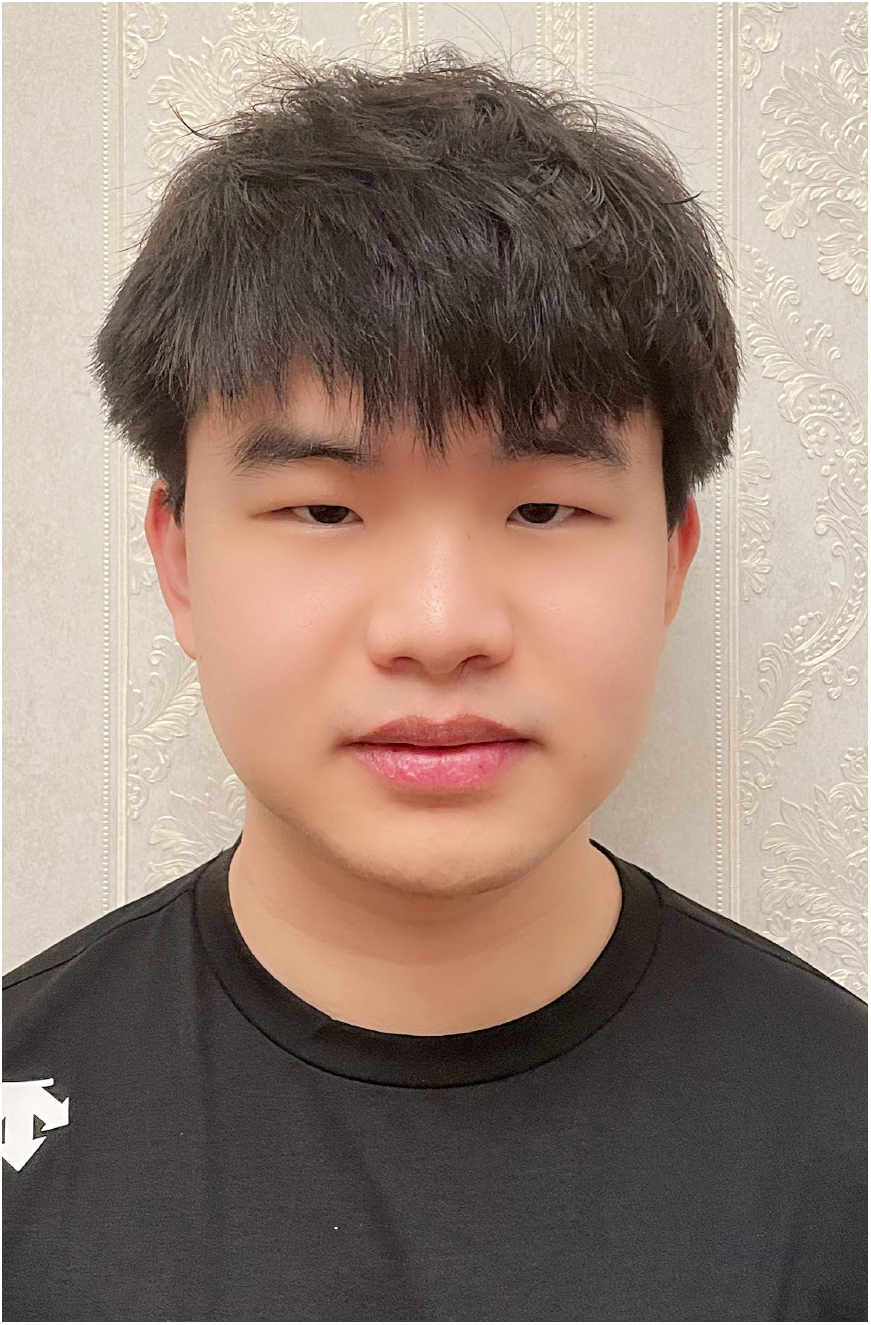}}]{Mudi Jiang}
	received the MS degree in software engineering from Dalian
University of Technology, China, in 2023. He is currently working toward
the PhD degree in the School of Software at the same university. His current
research interests include data mining and its applications.
\end{IEEEbiography}

\begin{IEEEbiography}[{\includegraphics[width=1in,height=1.25in,clip,keepaspectratio]{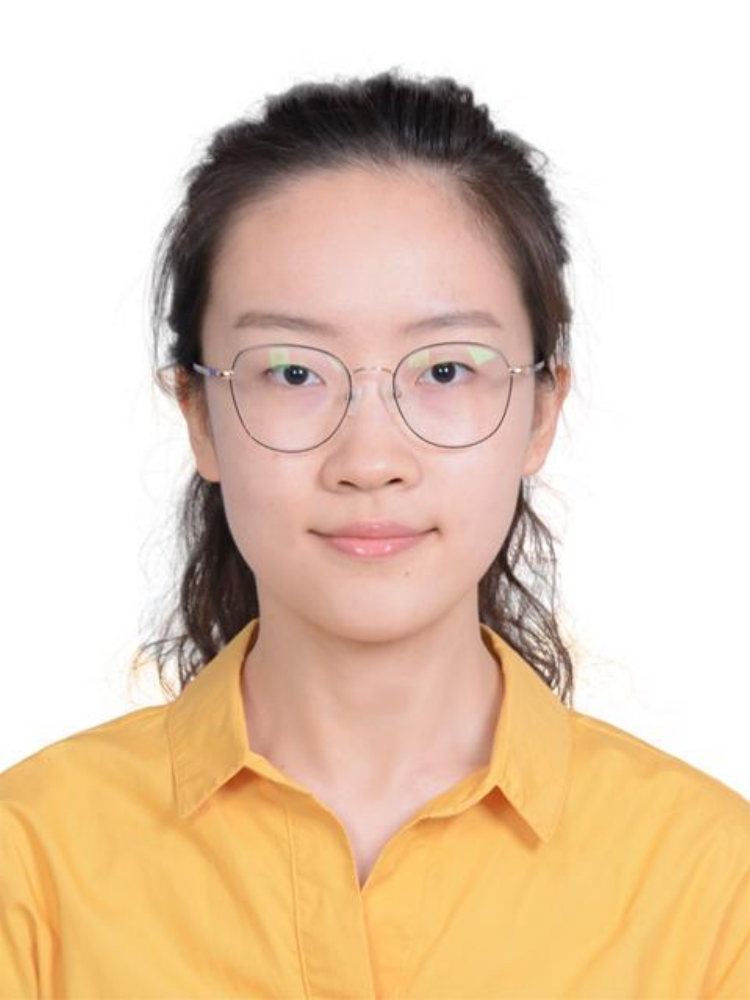}}]{Jiahui Zhou}
	received the BS degree from Dalian University of Technology, China, in 2022. She is currently working toward the MS degree in the School of Software at Dalian University of Technology. Her current
research interests include multimodal retrieval and data mining.
\end{IEEEbiography}

\begin{IEEEbiography}[{\includegraphics[width=1in,height=1.25in,clip,keepaspectratio]{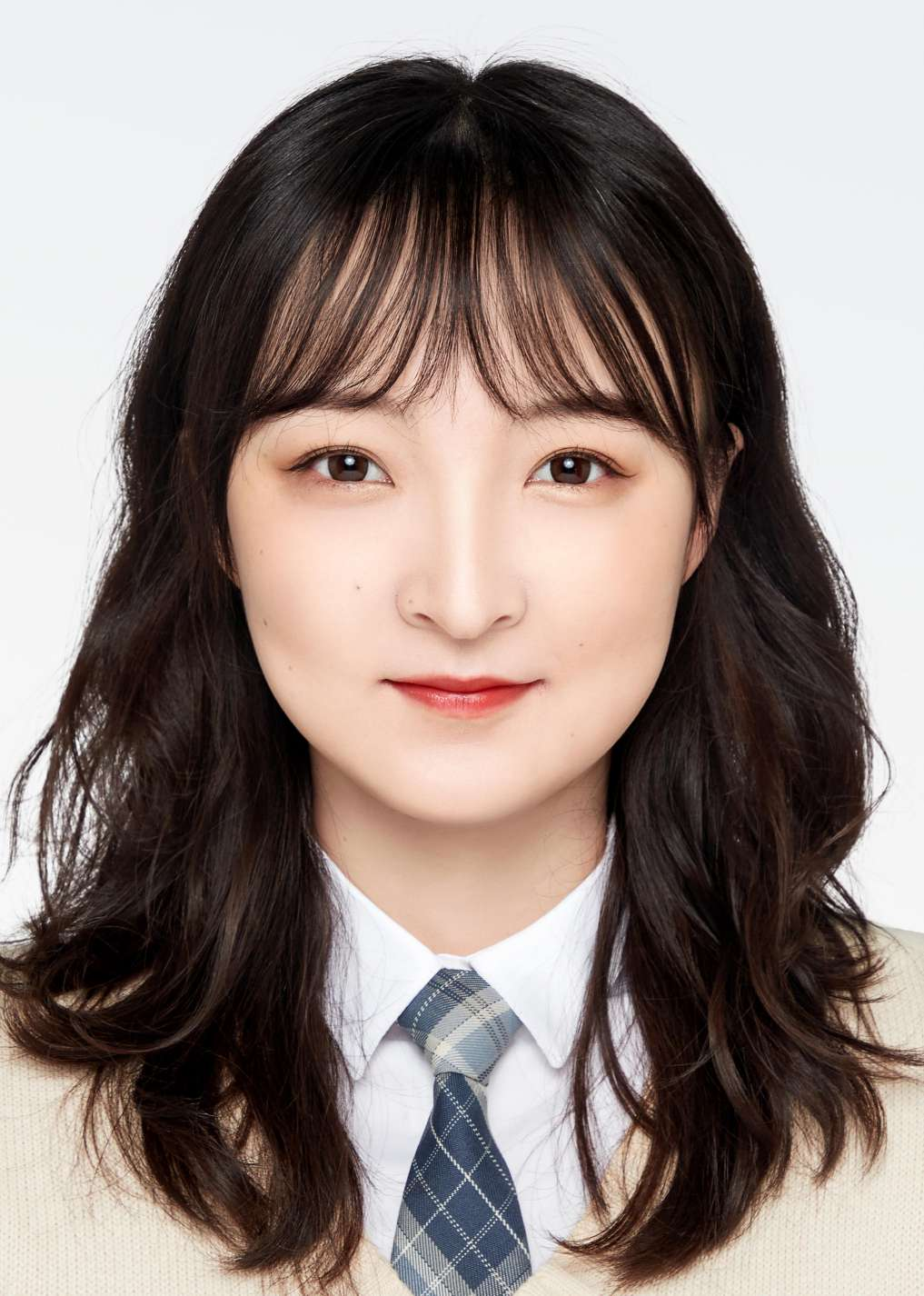}}]{Xinying Liu}
	received the MS degree in Applied Statistics from China University of Geosciences, China, in 2023. She is currently working toward the PhD degree in the School of Software at Dalian University of Technology. Her current research interests include machine learning and data mining.
\end{IEEEbiography}

\begin{IEEEbiography}[{\includegraphics[width=1in,height=1.25in,clip,keepaspectratio]{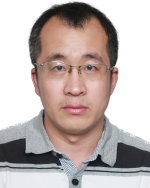}}]{Zengyou He}
	received the BS, MS, and PhD degrees in computer science from Harbin Institute of Technology, China, in 2000, 2002, and 2006, respectively. He was a research associate in the Department of Electronic and Computer Engineering, Hong Kong University of Science and Technology from February 2007 to February 2010. He is currently a professor in the School of software, Dalian University of Technology. His research interest include data mining and bioinformatics.
\end{IEEEbiography}

\begin{IEEEbiography}[{\includegraphics[width=1in,height=1.25in,clip,keepaspectratio]{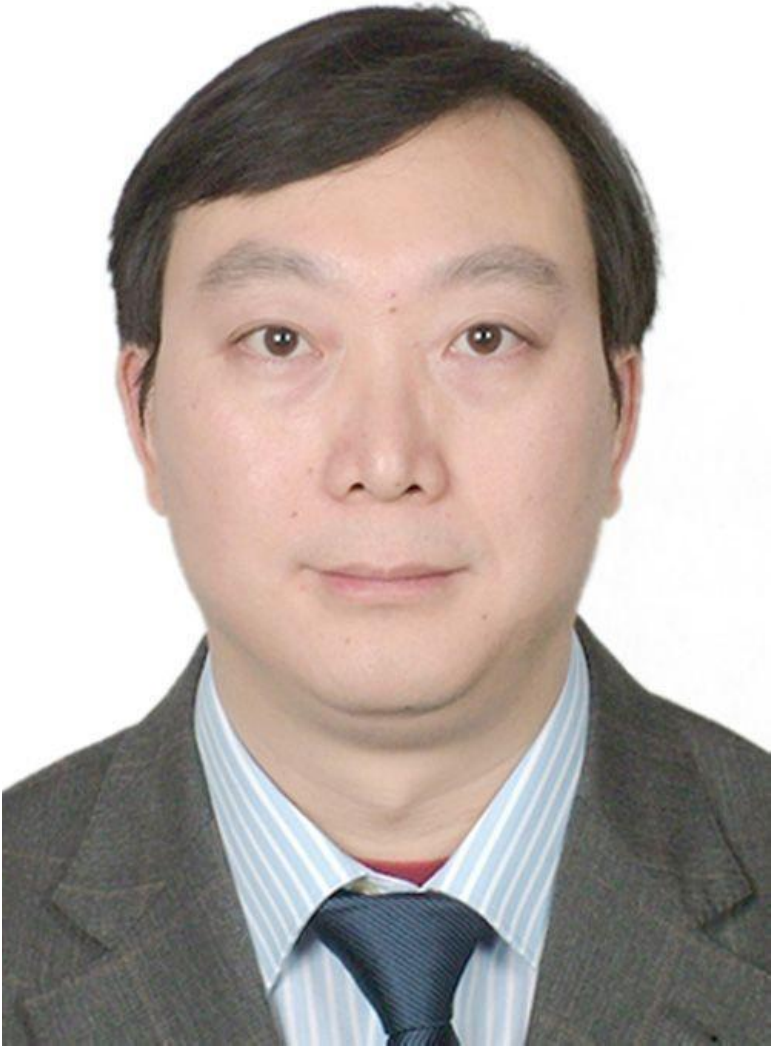}}]{Zhikui Chen}
	(Member, IEEE) received the B.S. degree in mathematics from Chongqing Normal University, Chongqing, China, in 1990, and the M.S. and Ph.D. degrees in mechanics from Chongqing University, Chongqing, in 1993 and 1998, respectively. He is currently a Full Professor with the Dalian University of Technology, Dalian, China.  His research interests are the Internet of Things, big data processing, mobile cloud computing, and ubiquitous networks.

\end{IEEEbiography}

\vfill

\end{document}